\def\year{2022}\relax
\documentclass[letterpaper]{article} 
\usepackage{aaai22}  
\usepackage{times}  
\usepackage{helvet}  
\usepackage{courier}  
\usepackage[hyphens]{url}  
\usepackage{graphicx} 
\usepackage{natbib}  
\usepackage{caption} 
\DeclareCaptionStyle{ruled}{labelfont=normalfont,labelsep=colon,strut=off} 
\usepackage{algorithm}
\usepackage{algorithmic}

\usepackage{newfloat}
\usepackage{listings}
\floatstyle{ruled}
\newfloat{listing}{tb}{lst}{}
\floatname{listing}{Listing}
\usepackage{todonotes}

\title{Signal Quality Auditing for Time-series Data}
\author{
    Chufan Gao\textsuperscript{\rm 1 \rm 2},
    Nicholas Gisolfi\textsuperscript{\rm 1},
    Artur Dubrawski\textsuperscript{\rm 1}
}
\affiliations{
    \textsuperscript{\rm 1} Carnegie Mellon University\\
    \textsuperscript{\rm 2} University of Illinois Urbana-Champaign\\

    chufang@andrew.cmu.edu, awd.cs.cmu.edu
}

\usepackage{bibentry}

\begin{document}

\maketitle

\begin{abstract}
Signal quality assessment (SQA) is required for monitoring the reliability of data acquisition systems, especially in AI-driven Predictive Maintenance (PMx) application contexts. 
SQA is vital for addressing "silent failures" of data acquisition hardware and software, which when unnoticed, misinform the users of data, creating the risk for incorrect decisions with unintended or even catastrophic consequences.
We have developed an open-source software implementation of signal quality indices (SQIs) for the analysis of time-series data.
We codify a range of SQIs, demonstrate them using established benchmark data, and show that they can be effective for signal quality assessment.
We also study alternative approaches to denoising time-series data in an attempt to improve the quality of the already degraded signal, and evaluate them empirically on relevant real-world data.
To our knowledge, our software toolkit is the first to provide an open source implementation of a broad range of signal quality assessment and improvement techniques validated on publicly available benchmark  data for ease of reproducibility. 
The generality of our framework can be easily extended to assessing reliability of arbitrary time-series measurements in complex systems, especially when morphological patterns of the waveform shapes and signal periodicity are of key interest in downstream analyses.

\end{abstract}









\section{Introduction} \label{secIntro}


Signal quality assessment (SQA) of time-series data is an important tool for monitoring reliability of data acquisition systems. 
It is vital for situational awareness of the maintainers and users of systems that rely on such data. 
Silent failures of data acquisition software and hardware may cause confusion in the consumers of the information being collected, and this confusion may then lead to untoward or even catastrophic consequences. 

The quality of an AI system is to a large extent a function of the quality of the data used to train data-driven models as well as the data quality of queries submitted for processing, and time-series data signals are often corrupted with domain-specific artifacts which can introduce structured noise that biases the learned decision logic of an AI system, while also distorting its inferences.
Examples of artifacts that often manifest in, e.g., healthcare time series data include baseline wander, muscle artifacts, powerline interference, or equipment failure \cite{satija2018review}.
These artifacts can lead to critical, potentially deadly, errors in clinical practice; similar negative outcomes can manifest when performing prognostic analyses of engineered systems.
Affected application areas include false alarm reduction \cite{hravnak2016real}, physiological pattern discovery \cite{gao2019detecting, gao2021identification}, determining cardiovascular sufficiency \cite{li2022automated, nagpal2019dynamically}, and more.
Issues with time-series data quality represent a major hurdle to the widespread integration of Artificial Intelligence into the PMx paradigm because poor data quality can significantly reduce the accuracy and applicability of machine learning algorithms. 

To illustrate our concepts, we focus on signal quality of clinically relevant data to leverage relatively rich prior work in that domain.
For instance, existing reference databases of electrocardiogram (ECG) data are excellent proxies for developing signal quality methods for broader application in the scenarios where the subjects of analyses are physical systems instead of human bodies.

While some work has been done in the area of open-source signal quality analysis, there are very few open source implementations of SQA tools for general use.
For example, Neurokit2\footnote{github.com/neuropsychology/NeuroKit}\cite{makowski2021neurokit2} - one of the largest ECG analysis libraries - only implements two signal quality indices, one of which does not work well in practice. 
Aura Healthcare\footnote{github.com/Aura-healthcare/ecg\_q} - another ECG analysis software repository - includes methods that overlap with Neurokit2. Neither of these collections offer benchmark results computed on publicly available reference data to enable their independent verification.  

We develop and analyze an open-source, reproducible, python implementation of many common SQA methods, including signal quality indices for analysis of electrocardiogram (ECG), plethysmography (Pleth), and other vital sign timeseries data prevalent in healthcare applications.
Additionally, we validate our results on both a closed-source bleeding dataset from a healthcare research organization
\cite{namas2009adequately}, and publicly available datasets such as the MIT-BIH Arrhythmia database~\cite{moody2001impact}.






The contributions of this work include:
\begin{itemize}
    \item Codifying core concepts in the benchmark analysis of Physionet 2011 ECG Quality Classification Challenge \cite{silva2011improving}. We show a new result--a combination of all methods yields better performance than any individual method alone.
    \item Showing that improvements to data by computing and leveraging SQIs can lead to improvements in downstream AI model performance. To demonstrate this, we consider a clinical alert adjudication task of determining whether the bedside monitor alert is due to real biological system status change, or to sensor artifacts.
    \item Considering SQIs in AI processing pipelines can reduce false detection and false negative rates in anomaly detection settings, when the goal is to identify subjects whose vital sign measurements look different from the rest of a cohort.
    \item Facilitating time-series signal denoising and present experimental results with both quantitative and qualitative improvements.
    \item Publicly releasing a software code repository for general use and reproduction of experimental results.
\end{itemize}

\section{Methodology and Experiments}



Descriptions of the methods, baselines, datasets and empirical results for four research questions relating to signal quality auditing for time-series data are presented here. \begin{enumerate}
    \item Which SQI methods yield top-performing models in signal quality classification?
    \item Which model classes benefit most from SQIs in an outlier detection task?
    \item How useful are SQIs at denoising time-series?
    \item What is the effect of ECG Denoising on Real vs
Artifactual Alert Detection?
\end{enumerate}

\subsection{Q1: Which SQI methods yield top-performing models in signal quality classification?}

\begin{table}[h]
\centering
\begin{tabular}{|l|l|}
\hline
\multicolumn{1}{|c|}{Method} & \multicolumn{1}{c|}{Signal Quality Indices (SQIs)} \\ \hline
(1) Li et al. 2007 & b\_sqi, p\_sqi, k\_sqi \\ \hline
(2) Clifford et al. 2012 & \begin{tabular}[c]{@{}l@{}}b\_sqi, p\_sqi, k\_sqi, s\_sqi, f\_sqi, \\ bas\_sqi\end{tabular} \\ \hline
(3) Behar et al. 2013 & k\_sqi, s\_sqi, p\_sqi, b\_sqi \\ \hline
(4) Li et al. 2014 & \begin{tabular}[c]{@{}l@{}}b\_sqi, p\_sqi, k\_sqi, s\_sqi, f\_sqi, \\ bas\_sqi, bs\_sqi, e\_sqi, hf\_sqi, \\ pur\_sqi, rsd\_sqi, ent\_sqi\end{tabular} \\ \hline
\begin{tabular}[c]{@{}l@{}}(5) Geometric \\ Features\end{tabular} & \begin{tabular}[c]{@{}l@{}}Median, IQR, and Slope of HR. \\ R-peak interval STD. Samp. entr., \\ appx. entr., Relative power, \\ and ratio of LF and HF bands\end{tabular} \\ \hline
\begin{tabular}[c]{@{}l@{}}(6) Average QRS \\ (Neurokit)\end{tabular} & averageQRS\_sqi \\ \hline
\begin{tabular}[c]{@{}l@{}}(7) Zhao et al. 2018 \\ (Neurokit)\end{tabular} & zhao2018\_sqi \\ \hline
\begin{tabular}[c]{@{}l@{}}(8) Orphanidou et al. \\ 2015\end{tabular} & orphanidou2015\_sqi \\ \hline
(9) Combined & \begin{tabular}[c]{@{}l@{}}All of the previously mentioned \\ SQIs and features\end{tabular} \\ \hline
\end{tabular}
\caption{We compare 9 methods shown in the table, with their corresponding SQI features. Methods (6), (7) , and (8) each yield a single SQI feature.}
\label{tab:sqi_methods}
\end{table}


We benchmarked a total of 8 prior works as well as a method that combines SQIs from all those 8 prior research threads.
\cite{li2008robust} was one of the first to apply machine learning to address ECG signal quality and introduced a few common ECG SQIs: a beat-agreement calculated from two beat-detection algorithms (bSQI), an inter-channel signal quality metric (iSQI), the kurtosis of the signal (kSQI), and the spectral distribution of ECG (pSQI).
\cite{clifford2012signal} added 3 more SQIs, including skewness SQI (sSQI), a flatness SQI (fSQI), and a feature to measure the relative power in the baseline (basSQI).
\cite{li2014machine} introduced 6 more SQIs: a baseline wander check in time domain (bsSQI), the relative energy in the QRS complex (eSQI), the relative amplitude of high frequency noise (hfSQI), the signal purity (purSQI), the relative standard deviation (STD) of QRS complex (rsdSQI), the sample entropy (entSQI), the high frequency mask of the ECG waveform (hfMSQI), and a periodic component analysis periodicity measure  (PiCASQI).
\cite{behar2013ecg} introduced an SQI based on the most significant principal components obtained by principal component analysis (pcaSQI).
Orphanidou et al. proposed a SQI based on average correlation between each beat and its average template. 
Neurokit2 \cite{makowski2021neurokit2} itself implemented an SQI that calculates a correlation of the QRS waveform to the average QRS waveform (averageqrsSQI).
Neurokit2 also implemented another method: Zhao et al. \cite{zhao2018sqi}, which used a combination qSQI, pSQI, kSQI, basSQ for a rule-based quality classification. Finally, we also compared geometric shape features calculated from ECG including: median, interquartile range (IQR), and the slope of heart rate (HR); r-peak interval standard deviation; sample entropy; approximate entropy; relative power; and ratio of low-frequency and high-frequency bands of the power spectrum of the ECG. The specific implementations of each method can be found in our code.

We applied these methods to the original Physionet 2011 ECG signal quality classification challenge, also known as the PICC dataset \cite{silva2011improving}.
This dataset included ten-second annotated recordings of twelve-lead ECGs, consisting of standard 12-lead ECG recordings, recorded simultaneously for a minimum of 10 seconds at 500 Hz.
ECG were annotated by nurses, technicians, and volunteers, where 3 to 17 annotators independently examined each ECG, to obtain a binary "noisy" or "not noisy" classification.
Additionally, this dataset was also used in an open classification challenge--
Table \ref{tab:picc_orig_results} shows the results of the challenge on the portion of the challenge that required open source code (the most relevant to our work).

\begin{table}[t]
\centering
\begin{tabular}{|ll|}
\hline
\multicolumn{1}{|l|}{\textbf{Participant}} & \textbf{Accuracy} \\ \hline
\multicolumn{1}{|l|}{Xiaopeng Zhao} & 0.914 \\ \hline
\multicolumn{1}{|l|}{Benjamin Moody} & 0.896 \\ \hline
\multicolumn{1}{|l|}{Lars Johannesen} & 0.880 \\ \hline
\multicolumn{1}{|l|}{Philip Langley} & 0.868 \\ \hline
\multicolumn{1}{|l|}{Dieter Hayn} & 0.834 \\ \hline
\multicolumn{1}{|l|}{Václav Chudáček} & 0.833 \\ \hline
\multicolumn{2}{|l|}{\textit{Unofficial entries}} \\ \hline
\multicolumn{1}{|l|}{George Moody} & 0.894 \\ \hline
\multicolumn{1}{|l|}{Ikaro Silva} & 0.802 \\ \hline
\end{tabular}
\caption{The PhysioNet/Computing in Cardiology Challenge 2011 Official Results: Event 2 (open-source, open data set B), the most comparable event to our case. However, this is still not exactly a proper comparison, since we do not have access to data Set B (only Set A).}
\label{tab:picc_orig_results}
\end{table}

We used 5-fold group split cross-validation on the subjects in Set-A of the dataset, so subject data do not overlap in training and testing in any iteration.
That yieldED 1996 ECG time-series for training and testing (50\% split), each 10 seconds long, recorded at 500 Hz.
These raw time-series signals are then featurized via the SQIs mentioned above.
This yieldED a straightforward binary classification: 0=not noisy, 1=noisy.
A Random Forest classifier was further trained on these SQI attributes.

\begin{table}[t]
\centering
\begin{tabular}{|l|l|l|}
\hline
Method & AUC & Accuracy \\ \hline
li2007 & 0.793 ± 0.033 & 0.869 ± 0.013 \\ \hline
clifford2012 & 0.829 ± 0.041 & 0.877 ± 0.014 \\ \hline
behar2013 & 0.803 ± 0.035 & 0.869 ± 0.013 \\ \hline
li2014 & 0.830 ± 0.041 & 0.882 ± 0.013 \\ \hline
orphanidou2015 & 0.826 ± 0.028 & 0.863 ± 0.012 \\ \hline
averageqrs & 0.700 ± 0.025 & 0.858 ± 0.013 \\ \hline
zhao2018 & 0.549 ± 0.013 & 0.792 ± 0.002 \\ \hline
geometric & 0.807 ± 0.028 & 0.875 ± 0.013 \\ \hline
all & \textbf{0.835 ± 0.033} & \textbf{0.888 ± 0.012} \\ \hline
\end{tabular}
\caption{Table of AUCs and Accuracies (with standard deviations) of 8 methods on subject-based 5-fold cross validation and Random Forest with \textit{only} single-lead features. A comination of all methods perform the best (bolded).}
\label{tab:sqi_single_lead_feats}
\end{table}

The results in Table \ref{tab:sqi_single_lead_feats} showed that using all SQI features is the best in both Area Under the Receiver Operating Characteristic (AUC) and classification accuracy. We also note that the features from \cite{li2014machine} performed close to all methods, indicating that their proposed SQIs (See Table \ref{tab:sqi_methods}) were highly informative in this task.
Additionally, we saw that, compared to the original PhysioNet/Computing in Cardiology Challenge 2011 results in Table \ref{tab:picc_orig_results}, the accuracy performance of the random forest ran on single-lead features was comparable to the 3rd spot on the leaderboard as shown in Table \ref{tab:sqi_single_lead_feats}, suggesting that we were able to reproduce baseline results. Furthermore, we saw that Neurokit2’s implementation of \cite{zhao2018sqi} does not fare well under actual evaluation. This highlights the effect of conducting evaluation on real-world benchmark data, and that using all features, including all SQIs and ECG features, appears to perform best.

\subsection{Q2: Which model classes benefit most from SQIs in an outlier detection task?}

In contrast to the previous question, here we considered the inherent ability of SQIs as outlier detection features, rather than explicit classification. Four popular Outlier Detection methods were evaluated for automatical detection of noisy signals.
First, K-Nearest Neighbor \cite{ramaswamy2000efficient} is a classic method of detecting outliers that is based on the distance of a point from its kth-nearest neighbor.
We can think of the outlier score as the distance to a sample’s kth-nearest neighbor.
Second, the Isolation Forest algorithm \cite{liu2008isolation} seeks to separate anomalous observations from other observations by computing the average number of random splits needed for an observation to reach leaf nodes of trees obtained by splitting data on randomly chosen features. 
The outlier score is the number of splittings required for a sample to reach a leaf, averaged over the forest.
A One-class Support Vector Machine (SVM) \cite{scholkopf2001estimating} can be thought of as a model that is fit to the data, where anomalous data is not well predicted by it.
Fourth, AutoEncoders (AEs) \cite{aggarwal2017introduction} are neural networks that map data onto low-dimensional embedding space, and then reconstruct data from the respective embedding.
Anomalies should yield higher reconstruction errors as the model is not trained on them as much as on regular data.

We trained each of these models on the PICC dataset as before, with the same 5-fold cross-validation strategy to ensure generalizability of our models. 
For implementation of these algorithms, we followed the example in the open-source, anomaly detection Python package PyOD\footnote{pyod.readthedocs.io}.
PyOD implemented 30 anomaly detection algorithms, with evaluation results on 55 benchmark datasets.
We instantiated each model as in the example code, trained on the train split of the PICC dataset, and then evaluated on the test split.
Each model is trained on a mix of clean and noisy signals, with the percentage of noisy signals in the dataset being known.
This is allows each model to find its own threshold for anomaly classification.
Finally, each model was tested to see if its predicted outliers match a "noisy" classification.

\begin{table}[H]
\centering
\begin{tabular}{|l|l|l|}
\hline
Method & Test AUC & Test Accuracy \\ \hline
KNN & 0.396 ± 0.013 & 0.693 ± 0.003 \\ \hline
IForest & \textbf{0.830 ± 0.006} & \textbf{0.844 ± 0.004} \\ \hline
OCSVM & 0.492 ± 0.062 & 0.646 ± 0.015 \\ \hline
AE & 0.819 ± 0.007 & 0.843 ± 0.003 \\ \hline
\end{tabular}
\caption{Results (with standard deviations) for the four outlier detection algorithms. IForest performs the best (bolded).}
\label{tab:od_test_models}
\end{table}

From the results in Table \ref{tab:od_test_models} we see that Autoencoders and Isolation Forest perform best in both metrics. In fact, the performance of Isolation Forest would have ranked 5th in the actual competition, despite the fact that it is only using information from the SQIs for outlier detection, not explicit classification. This indicates that the proposed SQIs are reflective of the true signal quality, which is promising for generalization to other signal quality assessment tasks.

\subsection{Q3: How useful  are SQIs at denoising time-series?}
\begin{figure}[t!]
    \centering
    \includegraphics[width=\columnwidth]{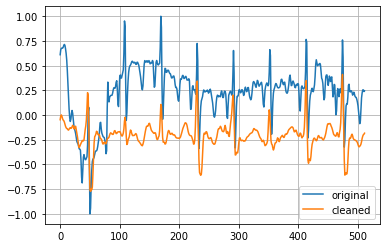}
    \includegraphics[width=\columnwidth]{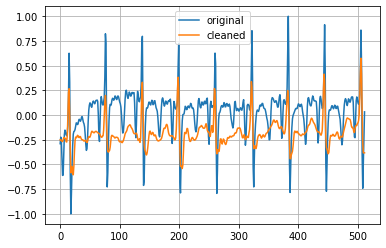}
    \includegraphics[width=\columnwidth]{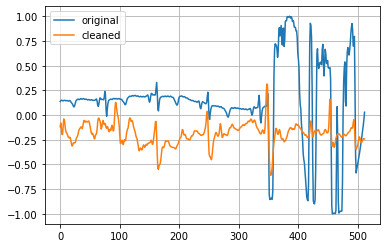}
    \caption{ECG Denoising Results example from the CNN autoencoder (the best model in terms of Mean Squared Error). The blue is the original signal, and the orange is the cleaned signal.}
    \label{fig:pig_dn}
\end{figure}

For this task, we tested three different methods for signal denoising at varying levels of signal quality. 
Wavelet denoising \cite{donoho1995noising} is an established method of denoising signals by discrete wavelet decomposition (DWT), thresholding the decomposed coefficients, and inverse wavelet decomposition (IDWT).
\cite{alfaouri2008ecg} presents an approach based on the threshold value of ECG signal determination using Wavelet Transform coefficients that applies thresholding on the decomposed coefficients.
For ECG denoising, it is common to use the Daubechies-4 (db4) wavelet.

Empirical Mode Decomposition (EMD) Denoising \cite{flandrin2004empirical} is a data-driven decomposition method that does not require any basis, unlike DWT.
\cite{weng2006ecg} proposed an ECG denoising method based on Empirical Mode Decomposition, which is able to remove high frequency noise with minimum signal distortion.
The goal of EMD is to decompose the signal into a sum of Intrinsic Mode Functions (IMFs), which are like DWT coefficients.
An IMF is defined as 1) function with equal number of extrema and zero crossings (or at most differed by one) and 2) The envelopes connecting the local minima and local maxima of an IMF have a local average of zero \cite{weng2006ecg}.
The process of extracting IMFs from a given input signal is called the "sifting process". 
CNN Denoising Autoencoders denoise signals by data compression and reconstruction.
\cite{chiang2019noise} introduced a wavelet denoising method using convolutional denoising autoencoders (AE) for ECG. 

\begin{table}[t]
\centering
\begin{tabular}{|c|c|c|c|c|}
\hline
\multicolumn{1}{|l|}{Noise} & dB & Wavelet MSE & EMD MSE & AE MSE \\ \hline
em & -6 & 0.0824 & 0.0684 & \textbf{0.0187} \\ \hline
em & 0 & 0.0527 & 0.0483 & \textbf{0.0104} \\ \hline
em & 6 & 0.0276 & 0.0337 & \textbf{0.0067} \\ \hline
em & 12 & 0.0101 & 0.0213 & \textbf{0.0055} \\ \hline
em & 18 & \textbf{0.0028} & 0.0171 & 0.0048 \\ \hline
em & 24 & \textbf{0.0008} & 0.0177 & 0.0046 \\ \hline
ma & -6 & 0.0839 & 0.0853 & \textbf{0.0109} \\ \hline
ma & 0 & 0.0612 & 0.0681 & \textbf{0.0076} \\ \hline
ma & 6 & 0.0249 & 0.0370 & \textbf{0.0063} \\ \hline
ma & 12 & 0.0097 & 0.0246 & \textbf{0.0054} \\ \hline
ma & 18 & \textbf{0.0026} & 0.0187 & 0.0047 \\ \hline
ma & 24 & \textbf{0.0010} & 0.0189 & 0.0046 \\ \hline
bw & -6 & 0.0794 & 0.0813 & \textbf{0.0063} \\ \hline
bw & 0 & 0.0608 & 0.0655 & \textbf{0.0060} \\ \hline
bw & 6 & 0.0378 & 0.0468 & \textbf{0.0057} \\ \hline
bw & 12 & 0.0178 & 0.0296 & \textbf{0.0057} \\ \hline
bw & 18 & 0.0077 & 0.0221 & \textbf{0.0045} \\ \hline
bw & 24 & \textbf{0.0012} & 0.0188 & 0.0051 \\ \hline
gn & -6 & 0.0885 & 0.0941 & \textbf{0.0272} \\ \hline
gn & 0 & 0.0437 & 0.0549 & \textbf{0.0144} \\ \hline
gn & 6 & 0.0221 & 0.0364 & \textbf{0.0066} \\ \hline
gn & 12 & 0.0066 & 0.0212 & \textbf{0.0054} \\ \hline
gn & 18 & \textbf{0.0026} & 0.0184 & 0.0047 \\ \hline
gn & 24 & \textbf{0.0011} & 0.0174 & 0.0044 \\ \hline
all & -6 & 0.0653 & 0.0685 & \textbf{0.0229} \\ \hline
all & 0 & 0.0487 & 0.0566 & \textbf{0.0105} \\ \hline
all & 6 & 0.0292 & 0.0412 & \textbf{0.0065} \\ \hline
all & 12 & 0.0113 & 0.0239 & \textbf{0.0055} \\ \hline
all & 18 & \textbf{0.0035} & 0.0187 & 0.0048 \\ \hline
all & 24 & \textbf{0.0011} & 0.0171 & 0.0045 \\ \hline
\end{tabular}
\caption{Mean Squared Error of the predicted output ECG vs the actual clean ECG for all 3 compared methods, at each type of noise and decibel. Bolded results indicate the best for that level and type of noise.}
\label{tab:denoising_mse}
\end{table}

\begin{figure*}[!t]
    \centering
    \includegraphics[trim={2.5cm 0 2.5cm 0},clip,width=\textwidth]{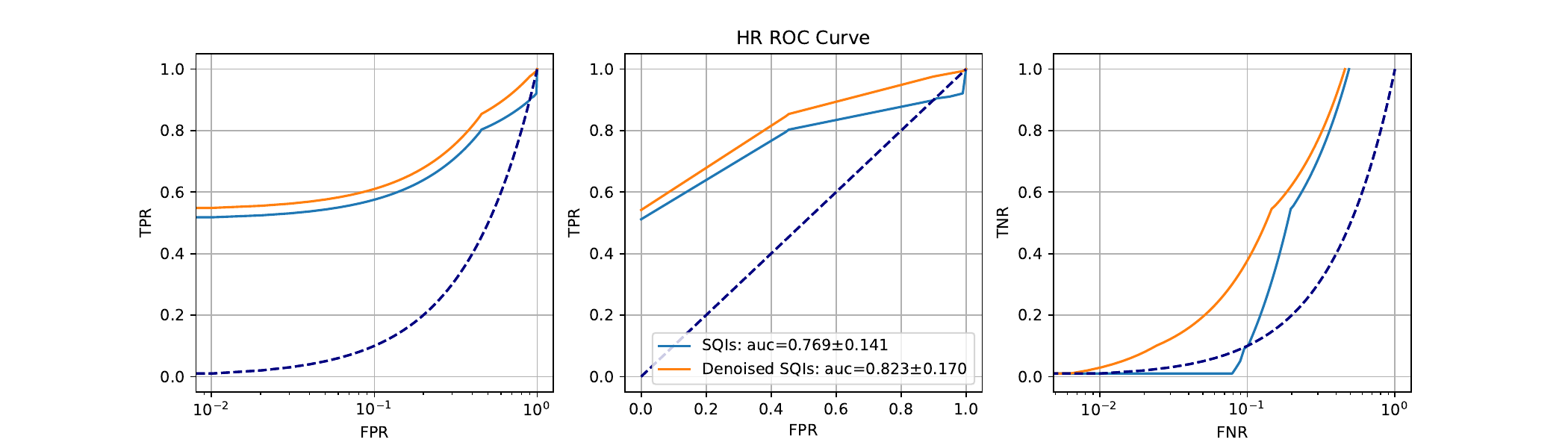}
    \caption{Plots of Receiver Operating Characteristic (ROC) curves (with AUC and stanrdard deviation) for Heart Rate alerts, with and without additional SQIs calculated on raw (represented by the blue line) and denoised ECG (represented by the orange line).}
    \label{fig:ppinnc_denoised}
\end{figure*}

We used 2 public datasets to test our denoising methods.
MIT-BIH arrhythmia and MIT-BIH arrhythmia noise stress test databases. These datasets will be used to construct ground truth noisy and clean ECG signals for evaluation.
The MIT-BIH Arrhythmia Database \cite{moody2001impact, mark1982annotated, moody1990bih} is an open access dataset available on PhysioNet \cite{goldberger2000physiobank} which contains 48 half-hour two-channel ambulatory ECG recordings, from 47 subjects.
The recordings were digitized at 360 hz per channel with 11-bit resolution over a 10 mV range.
Two or more cardiologists independently annotated almost every QRS signal.
Disagreements were resolved to obtain reference annotations for each beat (approximately 110,000 annotations total).
The MIT-BIH Noise Stress Test Database (NSTDB) \cite{moody1984noise} is another database available on PhysioNet \cite{goldberger2000physiobank} which includes 12 half-hour ECG recordings and 3 half-hour recordings of noise typical in ambulatory ECG recordings.
The three noise records were assembled from ECG signals that contained predominantly baseline wander (bw), muscle (EMG) artifact (ma), and electrode motion artifact (em) noise.

We manually added noise from the MIT-BIH dataset to control the type and variety of noise added.
For 2 clean records in the MIT-BIH dataset (118 and 119), we manually added 5 different types of noise (Electrode Motion artifact (EM), Baseline Wander (BW), Muscle Artifact (MA)), Gaussian Noise (GN), as well as a combination of all 4 previous noises at different levels of signal-to-noise ratio (SNR) of (-6, 0, 6, 12, 18, 24). SNR is a metric used to measure level of a desired signal to the level of background noise, where an SNR of 0 dB indicates that the signal level is equal to the noise level. The higher the ratio, the higher the signal level, and the better the signal quality.
All the noise was taken from the MIT-BIH noise stress test database except Gaussian Noise, which was manually added. Finally, we splitted ECG signals into windows of length 512, which is required by the autoencoder denoising method.
Splitting data into training and testing sets yielded 9408 samples for each and 18816 samples in total.

The full implementation details of the adapted methods are in the file \texttt{denoising.py} in the released python package.
The evaluation metric is Mean Squared Error of the reconstructed signal vs the ground truth signal.
For each method, the input will be a noisy signal, and the output is expected to be a cleaned version of that signal.


The results of the denoising is shown in Table \ref{tab:denoising_mse}. From the MSE table, we see that Convolutional Autoencoder generally performed the best at denoising ECG with low SNR (higher anounts of noise), whereas wavelet denoising performed better for higher SNR (lower amounts of noise). It would be interesting to investigate methods where wavelet denoising is used for higher SNR and Convolutional Autoencoders are used for lower SNR. 

Examples of noisy and denoised ECGs from the CNN autoencoder are shown in Figure \ref{fig:pig_dn}. From the figures, the cleaning process was able to at least remove some artifacts that occur in the original signal. For example, in the first figure, the baseline wander component was removed. Additionally, the CNN Autoencoder was able to preserve the peak to peak alignment quite well. 

However, there are still issues to be solved, as in the bottom right figure, in cases of extreme noise, the algorithm may not be able to accurately denoise the signal. The CNN Autoencoder looses information in the first 2/3 of the time window. This could be related to hyperparameter choice (kernel size, filters), and the strict requirement of the neural network for a fixed length input time series. The final section addresses future work to tackle and potentially solve these issues.

\subsection{Q4: What is the effect of ECG Denoising on Real vs Artifactual Alert Detection?}

For a real world predictive maintenance application, we tested the ability of our methods to improve robustness of CRI alert sysems by improving classification of real vs artifactual alert prediction. 
This is a binary classification task, Real Alert=1 and Artifactual Alert=0.
The CRI\_Alert dataset is not publicly available, but it is a good target for our anlysis.
It is a set of time series data from Intensive Care Unit patients from a healthcare research organization.
High rates of false alarms for cardiorespiratory instability (CRI) in monitored patients cause alarm fatigue.
The CRI\_Alert project was created to look at episodes of Cardio-Respiratory Insufficiency (CRI) with two primary objectives:
First, to distinguish between real and artifactual CRI alerts; Second, to predict when a patient is likely to have a CRI in the future.
In this dataset, we considered CRI alerts based on Heart Rate (HR). In terms of raw data, we have access to raw ECG time-series, which are split into 6537 3-min windows, every 20 seconds, corresponding to each alert. Since we have access to the raw ECG time-series signals, we were able to featurize the signal into our SQIs. The raw ECG signals were featurized by using all of the features as mentioned in Q1, as it performed the best in terms of signal quality classification, and should contains the most amount of information for the downstream task. This featurization created a total of 22 features (all listed SQIs excluding zhao2018\_sqi). Additionally, 10 patient level random splits were used for cross evaluation to ensure no overlap between patient data in the test-train splits.

A random Random Forest was trained on the SQIs from two considered cases of ECG signal. 
In the first case, the raw time-series was featurized as normal.
In the second case, a denoised version of the ECG was featurized instead.
This denoised version was obtained by resampling the raw ECG signal to a hz of 125 and processing it via the CNN Autoencoder. 

From Figure \ref{fig:ppinnc_denoised}, we see that the AUC performance of the Random Forest is significantly better compared to the results ran on Raw ECG signals for Heart Rate CRIs. This increase in performance indicates that the denoising algorithm is working, as it is removing noise from clean signals, and failing to remove noise from noisy ones, increasing the difference between the 2 classes. This effect helps the Random Forest model more efficiently distinguish between real and artifactual alerts, improving robustness and reducing downtime for clinicians.

\section{Discussion and Conclusion}

Data quality, and more specifically time-series signal quality, is a major concern in in all real-world application contexts where measurement uncertainty cannot be avoided.
This concern is particularly felt in AI-driven PMx for equipment, where the sensors responsible for measuring health of components are, themselves, components that may fail in unexpected ways.
To mitigate risk of negative modeling outcomes, we
studied common baseline methods in Signal Quality Indices, Outlier detection, and Signal Denesoising. 

Experiments that we performed dealt with clinically relevant data for convenience during the development phase, where we wanted to obtain copious amounts of publicly available data as well as numerous baselines in prior work to compare against.
This design choice was made due to challenges specific to AI-driven PMx contexts.
Many datasets are not publicly available, meaning that there is no opportunity to compare to meaningful baselines as we did with the PhysioNet Challenge.
Our methods are general enough that they can be applied in PMx contexts in a straightforward manner.

Our results show that SQIs are capable of reproducing 9 signal quality classification methods, showing that a combination of all methods performs the best in terms AUC and Accuracy.
We showed that an Isolation Forest is able to accurately determine whether a signal is noisy or not, showing that the SQIs are an accurate representation of the true signal quality.
In time-series signal denoising appliations, our results show CNN Autoencoders performed the best in terms of MSE when compared to two other state-of-the-art signal denoising methods.
Denoising significantly improved real-vs-artifact detection.
This demonstrates that there is a clear advantage in using SQIs as well as denoising in terms of a concrete downstream, predictive maintenance task of false alarm reduction. 
This conclusion in particular demonstrates the benefit that we stand to gain by focusing on improving data quality in times-series signals.

In terms of noisy signal classification, further work is required in terms of characterizing the fault modes of the system.
For example, investigating the types of noise that is specifically causing the time-series signal to create artifactual alerts would be vital in actively managing the reliability of measurement equipment. Additionally, expansions to more traditional applications of predictive maintenance, such as analyzing component vibration time-series to maintain aircraft, pumps, and more are a logical next step.

Future work may also focus on different design choices.
For example, the CNN Autoencoder denoising model may be further tuned in terms of kernel size, number of layers, and dimensionality.
These features have a large impact in model capability, so a more extensive evaluation of parameter choices is needed.
Grid search, Bayesian Optimization, or other techniques could be used to search for the best parameter choices for the network. 
Furthermore, the current model is limited with a fixed size requirement of the input signal; future methods should seek to address this, perhaps with alignment via Dynamic Time Warping, or by recent Neural Network methods that do not require a fixed length input.
Use of SQIs to measure signal quality in conjunction with the denoising model may also allow the user to quantify the amount of expected improvement.

We have released an open-source implementation (github.com/chufangao/signal\_quality) of all previously mentioned methods to serve as an accessible, open source, toolkit for signal quality analysis.
Our hope is that such a tool for quickly implementing baselines will facilitate more research in AI-driven PMx, specifically focusing on data quality auditing.

\bibliography{aaai22}

\end{document}